\documentclass{article} 
\usepackage[preprint]{colm2026_conference}

\usepackage[utf8]{inputenc}
\usepackage[T1]{fontenc}

\usepackage{microtype}
\usepackage{hyperref}
\usepackage{url}
\usepackage{booktabs}

\definecolor{darkblue}{rgb}{0, 0, 0.5}
\hypersetup{colorlinks=true, citecolor=darkblue, linkcolor=darkblue, urlcolor=darkblue}

\usepackage{amsmath}
\usepackage{amssymb}
\usepackage{mathtools}
\usepackage{amsthm}
\usepackage{multirow}
\usepackage{array}
\usepackage{xspace}
\usepackage{graphicx}
\usepackage{subcaption}
\usepackage[table]{xcolor}
\usepackage{pifont}
\usepackage{colortbl}
\usepackage{pgf}
\usepackage{caption}
\usepackage{float}
\usepackage{wrapfig}
\usepackage{enumitem}
\usepackage{listings}
\usepackage[most]{tcolorbox}
\usepackage{fontawesome}

\usepackage[capitalize,noabbrev]{cleveref}

\usepackage{adjustbox}

\theoremstyle{plain}

\theoremstyle{definition}

\theoremstyle{remark}

\newcommand{\eg}{\emph{e.g.,}\xspace}

\newcommand{\ignore}[1]{}

\definecolor{DeepBlue}{HTML}{2B3298}
\definecolor{LightBlueBg}{HTML}{F3F6FF}
\definecolor{DeepGreen}{HTML}{2E7D32}
\definecolor{LightGreenBg}{HTML}{F1F8E9}
\definecolor{DeepOrange}{HTML}{EF6C00}
\definecolor{LightOrangeBg}{HTML}{FFF3E0}
\definecolor{DeepPurple}{HTML}{7B1FA2}
\definecolor{LightPurpleBg}{HTML}{F3E5F5}
\definecolor{deepblue}{RGB}{0, 51, 153}

\lstdefinestyle{promptstyle}{
  basicstyle=\ttfamily\footnotesize,
  columns=fullflexible,
  breaklines=true,
  keepspaces=true,
  showstringspaces=false,
  upquote=true,
  frame=none,
  backgroundcolor={}
}

\newtcolorbox{PromptBox}[2][]{%
  enhanced, breakable,
  colframe=DeepBlue,
  colback=LightBlueBg,
  coltitle=white,
  boxrule=0.8pt, arc=2pt,
  fonttitle=\bfseries\large,
  title={#2},
  #1
}

\newcommand{\PromptSection}[2][DeepBlue]{%
  \par\medskip
  \begin{tcolorbox}[
    colback=#1!8!white,
    colframe=#1!0,
    left=4pt,right=4pt,top=2pt,bottom=2pt,
    arc=1mm, boxrule=0pt
  ]\centering\textbf{\textcolor{#1}{#2}}\end{tcolorbox}
  \vspace{-2pt}
}

\definecolor{darkgreen}{HTML}{008000}
\newcommand{\improvement}[2]{%
  \pgfmathparse{(#1 - #2)/#2 * 100}%
  \edef\impval{\pgfmathresult}%
  \textcolor{darkgreen}{\textbf{(+\pgfmathprintnumber[fixed, precision=1, fixed zerofill=true]{\impval}\%)}}
}

\title{Planner Matters! An Efficient and Unbalanced Multi-agent Collaboration Framework for Long-horizon Planning}

\author{Wenyi Wu\thanks{Equal contribution.} \And Sibo Zhu\footnotemark[1] \And Kun Zhou\thanks{Corresponding author: \href{mailto:franciskunzhou@gmail.com}{franciskunzhou@gmail.com}.} \And Biwei Huang \AND University of California, San Diego}

\begin{document}

\ifcolmpreprint
\fi

\maketitle

\begin{abstract}
Language model (LM)–based agents have demonstrated promising capabilities in automating complex tasks from natural language instructions, yet they continue to struggle with long-horizon planning and reasoning. 
To address this, we propose an enhanced multi-agent framework that decomposes automation into three roles: a planner for high-level decision-making, an actor for task execution, and a memory manager for contextual reasoning. While this modular decomposition aligns with established design patterns, our core contribution lies in a systematic compute-allocation analysis, revealing that planning is the dominant factor influencing task performance. Execution and memory management require significantly less compute and model capacity to achieve competitive results. Building on these insights, we introduce a planner-centric reinforcement learning approach, which exclusively optimizes the planner using trajectory-level rewards from a VLM-as-judge, while freezing the other components. 
Extensive experiments on benchmarks spanning web navigation, OS control, and tool use demonstrate that concentrating model capacity and learning on high-level planning yields robust and compute-efficient improvements in long-horizon agent automation.
Our code is publicly released.
\end{abstract}

{\centering
\begin{tabular}{ccc}
{\color{black}\faGithub}~\textbf{\href{https://github.com/WenyiWU0111/Planner-Matters-GUI-Agent}{Code}} &
{\color{deepblue}\faGlobe}~\textbf{\href{https://wenyiwu0111.github.io/Planner-Matters-project-page/}{Website}}
\end{tabular}
\par}
\vspace{0.5em}

\section{Introduction}

\begin{wrapfigure}{r}{0.55\textwidth}
    \centering
    \includegraphics[width=\linewidth]{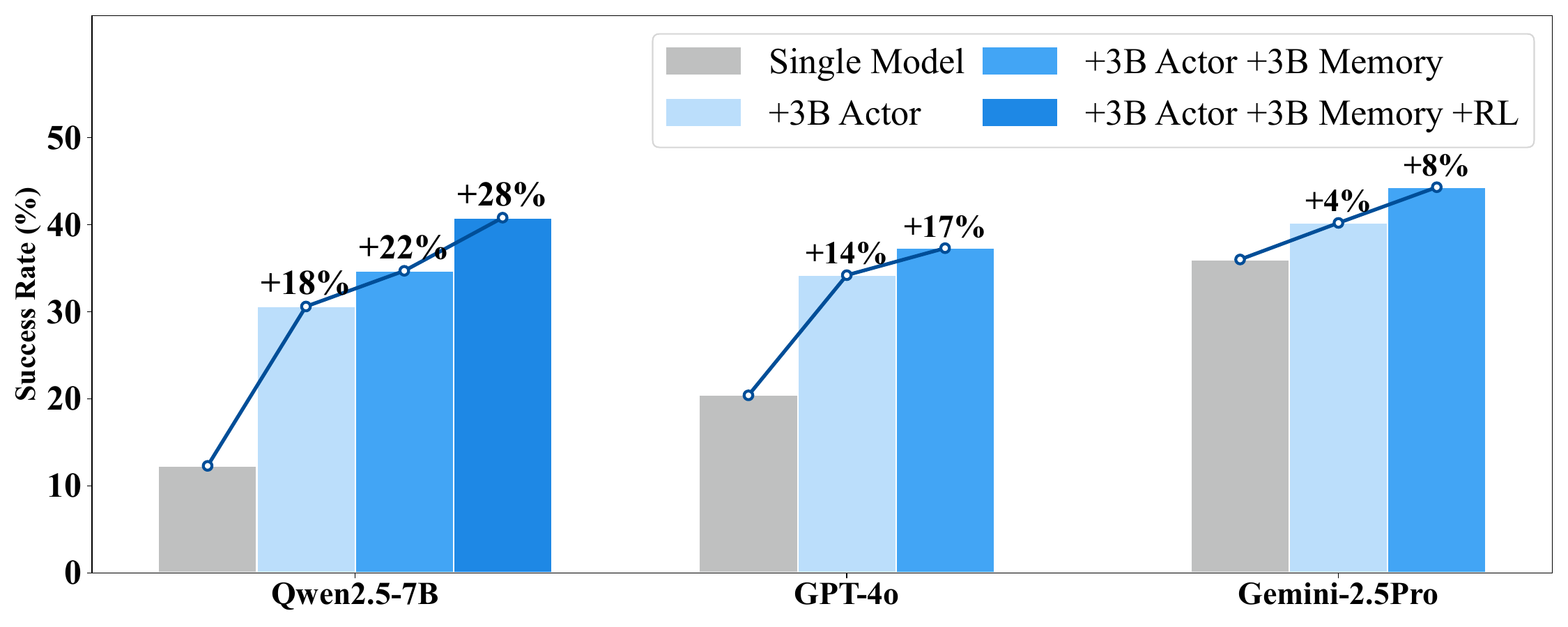}
    \caption{Performance on WebVoyager with planner-centered multi-agent setup across model scales. Using a strong planner with smaller actor and memory manager yields significant gains.}
    \label{fig:agent_comparison}
    \vspace{-1em}
\end{wrapfigure}

Driven by scaling laws, large vision–language models (VLMs)~\citep{bordes2024vlm_intro, zhang2024vlm_survey} have achieved remarkable performance across a wide range of multimodal reasoning and planning tasks. Building on this capability, agents powered by these models have demonstrated strong, human-like abilities in operating complex systems, enabling them to autonomously perform tasks such as web navigation, software control, and API-based tool use from natural language instructions~\citep{zhou2024webarenarealisticwebenvironment}. 
Typically, an agent interacts with a digital environment over multiple turns, jointly performing state interpretation, task planning, and action prediction to complete complex user-specified goals.

Despite these successes, existing agents still struggle with tasks that require long-horizon planning and reasoning, such as searching for complex compositional intents or conducting tasks that involve multi-source comparison and constrained filtering. A core difficulty is that it is hard for a single agent to simultaneously do well at both high-level goal management and low-level action execution. These tasks are challenging even for humans, because success requires maintaining the overall objective over time while also making accurate step-by-step decisions in a dynamic environment~\citep{cheng2025mgamemorydrivenguiagent}. 
When an agent focuses too much on low-level details, it may do the local steps right but still miss the overall goal, \eg click correctly, yet fail to compare options or apply constraints~\citep{erdogan2025planandactimprovingplanningagents, liang2026anticipatoryplanningmultimodalai}. Conversely, when an agent focuses more on the high-level objective, it may keep the overall right direction, but fails at execution through small mistakes, \eg missing a state change or skipping a check~\citep{tan2026enhancingwebagentshierarchical, aghzal2026llmbasedwebagentsfail}. 

To address the challenges of long-horizon planning, decomposing an agent's responsibilities into distinct yet cooperative components enables more structured and reliable decision-making~\citep{zhao2025colascalablemultiagentframework}. While prior work has explored planner--actor decompositions~\citep{erdogan2025planandactimprovingplanningagents,li2025agentflow}, a systematic understanding of \underline{\emph{how to allocate compute across components}} remains lacking. We adopt a multi-agent framework consisting of a planner, an actor, and a memory manager: the planner handles high-level goal decomposition; the memory manager maintains and updates relevant information; and the actor executes concrete actions. As shown in Figure~\ref{fig:agent_comparison}, this framework greatly improves performance, enabling Qwen2.5-VL-7B \footnote{All references to Qwen2.5-VL-7B indicate the Qwen2.5-VL-7B-Instruct model throughout this paper. The same applies to Qwen2.5-VL-32B, Qwen3-VL-4B, and Qwen3-VL-8B.} to outperform GPT-4o and Gemini.

Based on the effective multi-agent framework, we conduct a systematic \textbf{compute-allocation analysis} to study how individual components contribute to task success. Through controlled scaling experiments, we identify that the planner is the primary bottleneck, whereas actor and memory components require significantly reduced computational capacity to achieve comparable performance.
Our key findings reveal that planning capacity dominates overall performance. Scaling coefficients suggest that upgrading the planner alone achieves a performance gain comparable to scaling all components simultaneously, while the actor and memory manager have substantially lower contributions. This insight aligns with principles in cognitive neuroscience, where high-level executive control is recognized as a critical bottleneck in complex real-world tasks~\citep{miller2001integrative, szczepanski2014lesions}.

Building on this insight, we propose a \textbf{planner-centric reinforcement learning} (RL) approach. It exclusively optimizes the planner using trajectory-level rewards derived from VLM-as-judge, while keeping the actor and memory manager frozen. This targeted optimization strategy significantly improves agent performance. For example, with an unbalanced 7B+3B+3B configuration, our framework increases performance by 28\%, while achieving results comparable to closed-source baselines including GPT-4o, Gemini-2.5-Pro and Claude-4. These findings highlight the importance of disentangling and prioritizing planning capacity to efficiently achieve state-of-the-art performance in long-horizon agents.

\section{Preliminary}
\label{sec:prelim}

\paragraph{Long-horizon agent task.}
We address the problem of long-horizon task automation, where an agent interacts with a digital environment to execute complex user-specified tasks.
Each episode initializes with a natural-language task description $\mathcal{Q}$ (\eg ``book a flight'' or ``find a recipe and save it'').
At each time step $t$, the environment yields an observation $\mathcal{O}_t$, typically comprising a high-resolution screenshot and structured UI signals (\eg a DOM or accessibility tree).
The agent generates an action $\mathcal{A}_t \in \mathcal{A}$ for execution.

\paragraph{Action space.}
The action space $\mathcal{A}$ consists of parameterized interaction primitives. Each action is a tuple $(\texttt{type}, \texttt{params})$ where $\texttt{type} \in \{\texttt{Click}, \texttt{Type}, \texttt{Scroll}, \texttt{Select}, \texttt{Stop}, \texttt{ToolInvoke}\}$. Actions for GUI tasks are executed using Playwright\footnote{\url{https://playwright.dev/} 
} or PyAutoGUI\footnote{\url{https://pyautogui.readthedocs.io/} 
}, depending on the nature of the task. It enables the agent to interact with digital environments efficiently, while actions for tool-use tasks involve invoking specific tool APIs with corresponding parameters. The episode concludes upon task success, failure, or when a maximum step limit is reached. A formal definition of each action type and its parameters is provided in Appendix~\ref{append:action_space}.

\paragraph{Multi-agent systems.}
Unlike a monolithic agent that uses a single policy $\pi(\mathcal{A}_t \mid \mathcal{O}_t, \mathcal{Q})$, a multi-agent system decomposes decision-making into specialized modules. Each agent handles a distinct sub-capability, such as reasoning, memory retrieval, or action execution, and operates within its own local context. The global action $\mathcal{A}_t$ is produced through structured message passing where agents exchange intermediate outputs $z_t$ (\eg natural language plans, subgoals, or memory queries). This effectively separates high-level cognitive processes (planning) from low-level control (execution).

\begin{figure*}[t]
    \centering
    \includegraphics[width=0.9\textwidth]{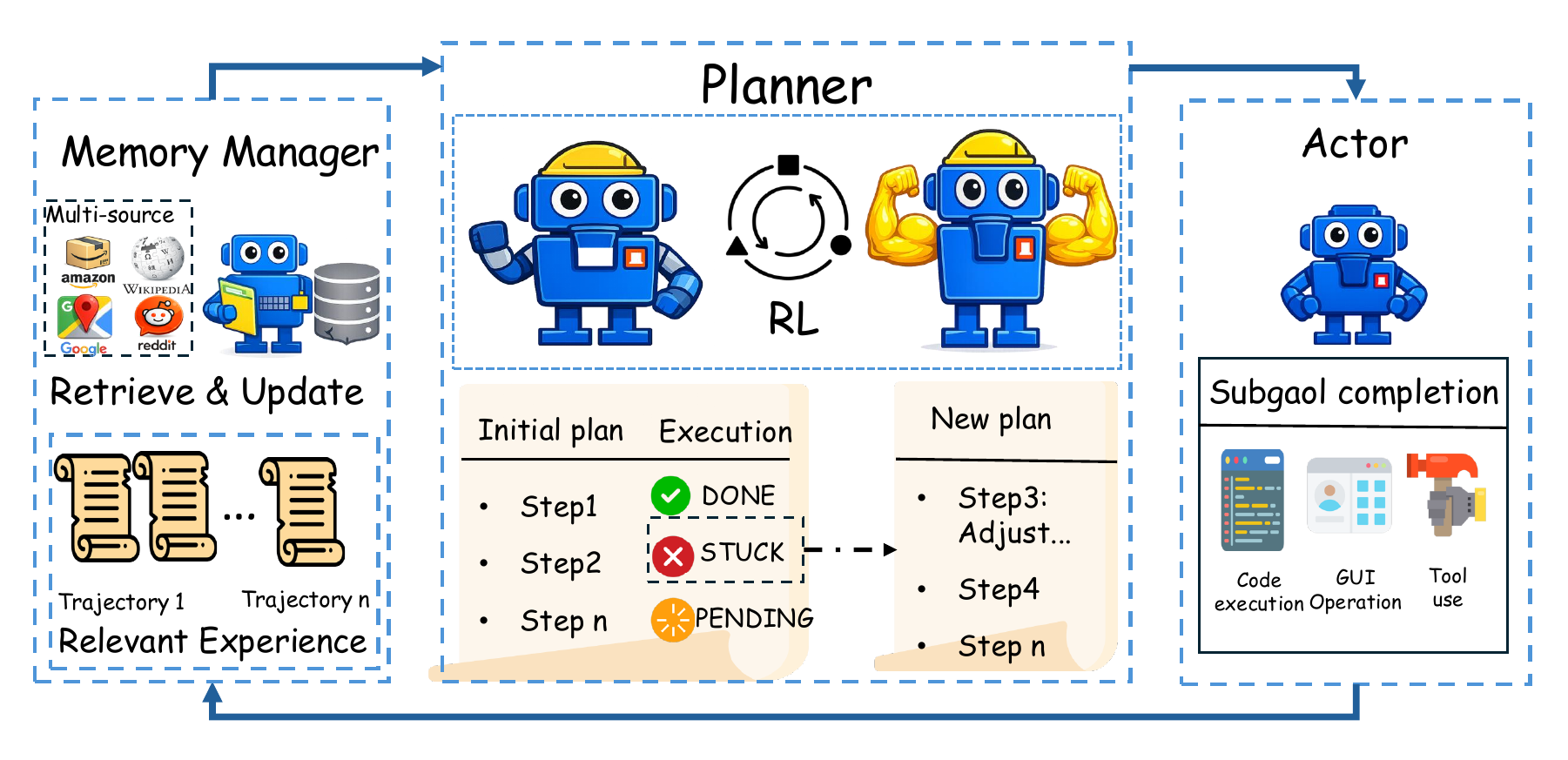}
    \vspace{-5pt}
    \caption{\textbf{Overview of the planner-centric multi-agent framework.} The system decomposes tasks into three agents: a central Planner that reasons over long-horizon goals; a Memory Manager that retrieves relevant discrete and continuous memories; and an Actor that executes precise actions. The pipeline supports optimization via reinforcement learning.}
    \label{fig:framework_arch}
\end{figure*}

\section{Multi-agent framework}
\label{sec:multi-agent-method}

\subsection{Multi-agent collaboration}

We propose a planner-centric multi-agent system consisting of three components: a \textit{Planner} as the central decision-maker, an \textit{Actor} that executes commands, and a \textit{Memory Manager} that provides dynamic context. These interact in an iterative loop throughout the task.

\paragraph{Planner.}
The Planner serves as the central decision-maker. At each step, it receives the task description $\mathcal{Q}$, the current observation $\mathcal{O}_t$, and the memory $\mathcal{M}_t$ from the Memory Manager.
It generates an initial plan at the beginning and incrementally updates it in subsequent steps based on the latest observations, the Actor's execution history, and updated memory. At each turn, it communicates the plan and focused subgoal to the Actor:
\begin{equation}
\mathcal{P}_t = \text{Planner}(\mathcal{Q}, \mathcal{O}_t, \mathcal{M}_t)
\end{equation}
where $\mathcal{P}_t$ denotes the current plan and focused subgoal to be executed.

\paragraph{Actor.}
The Actor translates the Planner's subgoal into a concrete, executable action $\mathcal{A}_t$. It takes as input the task description $\mathcal{Q}$, the Planner's plan $\mathcal{P}_t$, the current observation $\mathcal{O}_t$, and the available action space $\mathcal{A}$:
\begin{equation}
\mathcal{A}_t = \text{Actor}(\mathcal{Q}, \mathcal{P}_t, \mathcal{O}_t, \mathcal{A})
\end{equation}
After execution, the Actor returns the updated observation $\mathcal{O}_{t+1}$ to the system.

\paragraph{Memory manager.}
The Memory Manager retrieves and manages relevant memory to support the Planner. It processes the task description $\mathcal{Q}$, recent observations $\mathcal{O}_{t-k:t}$, and execution history $\mathcal{A}_{t-k:t}$ to provide the Planner with hybrid memory ~\citep{zhu2026hybridselfevolvingstructuredmemory} $\mathcal{M}_t$, which includes explicit discrete memory $\mathcal{M}_t^d$ and implicit continuous memory $\mathcal{M}_t^c$.
The discrete memory $\mathcal{M}_t^d$ consists of structured, text-based summaries of critical steps from past successful trajectories.
The continuous memory $\mathcal{M}_t^c$ consists of latent representations that encode long-term, multimodal trajectory information into compact embeddings~\citep{wu2025comem}.
\begin{equation}
\mathcal{M}_t = \text{MemoryManager}(\mathcal{Q}, \mathcal{O}_{t-k:t}, \mathcal{A}_{t-k:t})
\end{equation}
The Memory Manager also decides when to update memory. Let $\delta_t \in \{0, 1\}$ denote the update decision.
The next memory state is:
\begin{equation}
\mathcal{M}_{t+1} = (1 - \delta_t) \cdot \mathcal{M}_t + \delta_t \cdot \mathcal{M}_t';~\text{where}~\mathcal{M}_t' = \text{UpdateMemory}(\mathcal{M}_t, \mathcal{O}_{t+1}, \mathcal{A}_t).
\end{equation}
Details about memory construction can be found in Appendix~\ref{append:memory_construct}.

\begin{table*}[t]
\centering
\caption{Performance comparison of multi-agent configurations across different VLMs on four WebVoyager domains.}
\label{tab:decomposition_comparison}
\renewcommand{\arraystretch}{1}
\setlength{\tabcolsep}{2pt}
\small
\begin{tabular}{llccccc}
\toprule
\textbf{Model} & \textbf{Setting} & \textbf{Amazon} & \textbf{Google Map} & \textbf{Coursera} & \textbf{Allrecipes} & \textbf{Overall} \\
\midrule
\multirow{3}{*}{Qwen2.5-VL-7B}
& Single Model               & 14.6\% & 16.7\% & 2.4\%  & 15.6\% & 12.3\% \\
&  +Planner+Actor          & 53.7\% & 43.9\% & \textbf{38.1\%} & 26.7\% & 40.6\% \\
& +Planner+Actor+Memory & \textbf{56.1\%} & \textbf{53.7\%} & 33.3\% & \textbf{35.6\%} & \textbf{44.7\%} \\
\midrule
\multirow{3}{*}{GPT-4o}
& Single Model               & 24.4\% & 36.6\% & 7.1\%  & 13.3\% & 20.4\% \\
& +Planner+Actor          & 53.7\% & \textbf{51.7\%} & 35.7\% & \textbf{35.7\%} & 44.2\% \\
& +Planner+Actor+Memory & \textbf{55.6\%} & \textbf{51.7\%} & \textbf{47.6\%} & 34.1\% & \textbf{47.3\%} \\
\midrule
\multirow{3}{*}{Gemini-2.5-Pro}
& Single Model               & 41.7\% & 53.7\% & 28.6\% & 20.0\% & 36.0\% \\
& +Planner+Actor          & \textbf{51.2\%} & 56.1\% & 31.0\% & 24.4\% & 40.7\% \\
& +Planner+Actor+Memory & \textbf{51.2\%} & \textbf{58.5\%} & \textbf{35.7\%} & \textbf{28.9\%} & \textbf{43.6\%} \\
\bottomrule
\end{tabular}
\end{table*}

\subsection{Empirical analysis}

\paragraph{Effectiveness of multi-agent decomposition.}
To assess the necessity and effectiveness of our multi-agent design, we compare a monolithic model baseline with two variants: one using Planner and Actor agents, and another further adding a Memory Manager. Experiments are conducted with Qwen2.5-VL-7B-Instruct~\citep{bai2025qwen25vltechnicalreport}, GPT-4o~\citep{openai2024gpt4o}, and Gemini-2.5-Pro~\citep{team2023gemini} on four WebVoyager~\citep{he2024webvoyager} domains. All closed-source baselines use a single-model, screenshot-only setup with ReAct-style prompting and no external tool access.
As shown in Table~\ref{tab:decomposition_comparison}, introducing multi-agent decomposition significantly improves performance across all models: GPT-4o improves from 20.4\% to 44.2\%, and Qwen2.5-VL-7B from 12.3\% to 40.6\%. Adding memory further boosts results: GPT-4o's Coursera score rises from 35.7\% to 47.6\%. These gains confirm that decomposition is effective and generalizable.


\begin{figure*}[t]
    \centering
    \includegraphics[width=1\linewidth]{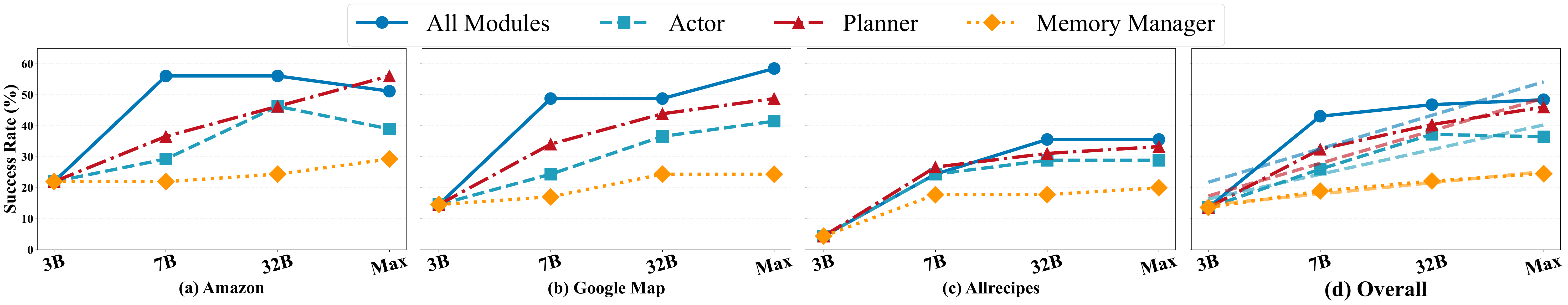}
    \caption{Performance when scaling individual agents (Actor, Planner, Memory Manager), with ``Max'' denoting Gemini-2.5-Pro. Scaling the Planner yields the largest marginal gains.}
    \label{fig:scal_module}
\end{figure*}

\paragraph{Scaling analysis: the planner matters most.}
To understand where model capacity should be allocated, we conduct a controlled ablation that selectively scales one component at a time while keeping others fixed. We evaluate across three domains (Amazon, Google Maps, Allrecipes) using Qwen2.5-VL-3B/7B/32B and Gemini-2.5-Pro as backbones.
As shown in Figure~\ref{fig:scal_module}, scaling the Planner yields the largest marginal performance gains compared to scaling either the Actor or Memory Manager individually. While the Actor also benefits from scaling, the Planner consistently captures most of the improvement achievable by scaling all modules simultaneously. The Memory Manager shows the smallest sensitivity to scale, suggesting that memory retrieval and management can be effectively handled by smaller models. This trend holds across models and domains.

\begin{wraptable}{r}{0.45\textwidth}
\centering
\caption{Fitted scaling law coefficients.}
\label{tab:scaling_law}
\small
\begin{tabular}{lccc}
\toprule
\textbf{Component} & \boldmath$\alpha$ & \boldmath$\beta$ & \boldmath$R^2$ \\
\midrule
All Modules       & 15.6 & 18.1 & 0.58 \\
Planner           & 16.0 & 12.7 & 0.82 \\
Actor             & 12.0 & 13.0 & 0.76 \\
Memory Manager    & 5.6 & 12.7 & 0.89 \\
\bottomrule
\end{tabular}
\end{wraptable}

\paragraph{Scaling law for parameter allocation.}
To quantify scaling behavior, we fit a linear regression $y = \alpha \log(x) + \beta$ where $x$ is model size (in billions of parameters)\footnote{We estimate the parameter size of GPT-4o as 200B.} and $y$ is the overall success rate (\%). Fitted lines are plotted in Figure~\ref{fig:scal_module}(d), and Table~\ref{tab:scaling_law} summarizes the coefficients.
The Planner achieves $\alpha=16.0$, nearly matching the all-modules coefficient ($\alpha=15.6$), confirming that most performance gains come from increasing Planner capacity. The Actor ($\alpha=12.0$) also benefits from scale but contributes less, while the Memory Manager ($\alpha=5.6$) shows the weakest scaling response. This quantitative analysis provides practical guidance for compute allocation under constrained budgets.

\section{RL for improving multi-agent collaboration}

Despite the improvements from decomposition, the Planner can still make critical mistakes: suboptimal goal setting, overly aggressive decisions, or premature termination (illustrated in Appendix~\ref{append:case_study_planner}).
Long-horizon decision-making and subgoal decomposition are difficult to optimize through supervised learning alone, especially in dynamic, partially observable environments~\citep{luo2025guir1}. We therefore turn to reinforcement learning to improve the Planner's ability to learn from sequential feedback, optimize for long-term rewards, and adapt strategies through exploration.
We note that this is best understood as \emph{modular single-policy RL}, we only optimize one module within a frozen pipeline, rather than multi-agent RL in the traditional cooperative/competitive sense.

\subsection{RL with trajectory-level rewards}
\label{sec:appro_rl}

We apply Group Relative Policy Optimization (GRPO)~\citep{shao2024deepseekmath} to fine-tune the Planner in an online, interactive setting.

\paragraph{Trajectory-level reward via VLM evaluation.}
To provide stable feedback for long-horizon planning and avoid brittle credit assignment to intermediate steps, we adopt a final-outcome-based reward. After execution, a VLM evaluator (Qwen2.5-VL-32B) assesses the full episode: plans, screenshots, and final answer, and assigns a score $r_i \in \{1, 3, 5\}$ based on criteria including task correctness, reasoning coherence, and interaction efficiency. The full evaluation prompt and criteria are provided in Appendix~\ref{append:judge_prompt}.
To mitigate noise, we use a voting mechanism across $K$ independent evaluations:
$r = \text{mode} \left( \{ r_i \}_{i=1}^K \right)$.
This trajectory-level reward is broadcast to all planning actions.

\paragraph{Algorithm.}
We optimize the Planner using GRPO with on-policy rollouts. With $(\mathcal{Q}, \mathcal{O}_t, \mathcal{M}_t, \mathcal{P}_t) \sim \mathcal{E}$ as experiences sampled from rollouts $\mathcal{E}$, the objective is:
\begin{equation}
J(\theta) = \mathbb{E}_{(\mathcal{Q}, \mathcal{O}_t, \mathcal{M}_t, \mathcal{P}_t) \sim \mathcal{E}} \left[ \rho_t A_t - \beta D_{\mathrm{KL}}[\pi_\theta \| \pi_{\text{ref}}] \right];~\text{where}~\rho_t = \frac{\pi_\theta(\mathcal{P}_t \mid \mathcal{Q}, \mathcal{O}_t, \mathcal{M}_t)}{\pi_{\theta_{\text{old}}}(\mathcal{P}_t \mid \mathcal{Q}, \mathcal{O}_t, \mathcal{M}_t)}
\end{equation}
where $\rho$ is the importance sampling ratio, $D_{\mathrm{KL}}[\pi_\theta \| \pi_{\text{ref}}]$ is the KL divergence from a reference policy, and $\beta$ is the regularization coefficient.
Since the reward is computed at the trajectory level, we assign the same score to all steps within a rollout and normalize across the group. The group-normalized advantage is:
\begin{equation}
\small
A_i^t = \frac{
    r(\mathbf{e}_i) - \mathrm{mean} \left( \left\{ r(\mathbf{e}_k) \right\}_{k=1}^G \right)
}{
    \mathrm{std} \left( \left\{ r(\mathbf{e}_k) \right\}_{k=1}^G \right)
}
\end{equation}
where $\mathbf{e}_i = (\mathcal{Q}, \mathcal{O}_i, \mathcal{M}_i, \mathcal{P}_i)$ denotes the $i$-th experience. We acknowledge that broadcasting trajectory-level rewards to all steps makes credit assignment coarse; incorporating step-level process rewards~\citep{chae2025webshepherdadvancingprmsreinforcing} is a promising direction for future work.

\subsection{Planner-only efficient training}

We freeze the Actor and Memory Manager during training, ensuring training focuses solely on enhancing the Planner's ability to generate high-quality plans. During each episode, the Memory Manager provides retrieved memory, the Planner generates a plan and subgoals, and the frozen Actor executes. After the full trajectory is collected, a trajectory-level reward is assigned and used to update the Planner's policy.

\section{Experiments}

\subsection{Experimental setup}

\paragraph{Memory settings.}
We construct the memory bank from successful trajectories generated by our agent on websites including Amazon, Google Maps, Coursera, and Allrecipes, domains covered by WebVoyager~\citep{he2024webvoyager}. These rollouts serve as the source for both discrete and continuous memory. We set the initial number of retrieved examples to 10 and enable dynamic memory update during evaluation. Memory construction details and dynamic memory update mechanism can be found in Appendices~\ref{append:task_generation} and~\ref{append:training_params}.

\paragraph{Training settings.}
We train all models using online GRPO~\citep{shao2024deepseekmath} with trajectory-level rewards from Qwen2.5-VL-32B via voting. Training is conducted on 320 tasks across four real-world websites, which were generated through a semi-automated pipeline. It ensures sufficient executability, diversity, and reliability for training tasks. Details about task generation, filtering, and hyperparameters are in Appendices~\ref{append:task_generation} and~\ref{append:training_params}.

\paragraph{Evaluation tasks.}
We evaluate on three web-agent benchmarks: WebVoyager~\citep{he2024webvoyager} (in-domain), Mind2Web~\citep{deng2023mind2webgeneralistagentweb} and MMInA~\citep{tian2025mminabenchmarkingmultihopmultimodal} (out-of-domain). We report task accuracy as our primary metric, evaluated with standard protocols. Additionally, we evaluate on OSWorld~\citep{xie2024osworldbenchmarkingmultimodalagents} for OS-level tasks and MCPBench~\citep{wang2025mcpbench} for tool-use tasks.

\subsection{Baseline methods}

\paragraph{Closed-source base models.}
We compare against GPT-4o~\citep{openai2024gpt4o}, Gemini-2.5-Pro~\citep{team2023gemini}, and Claude-4~\citep{anthropic2025claude}. All use a single-model setup with screenshot-only observation and ReAct-style prompting, without access to external tools or agentic scaffolding. We note that systems with full agentic scaffolding (\eg OpenAI Operator~\citep{openai2025operator}) achieve substantially higher absolute numbers.

\paragraph{Open-source base models.}
We include Qwen2.5-VL-32B~\citep{bai2025qwen25vltechnicalreport}, CogAgent~\citep{hong2024cogagentvisuallanguagemodel}, WebSight~\citep{bhathal2025websightvisionfirstarchitecturerobust}, and UI-TARS-1.5-7B~\citep{qin2025uitarspioneeringautomatedgui}.

\paragraph{Memory-augmented baselines.}
ReasoningBank~\citep{ouyang2025reasoningbankscalingagentselfevolving} and Agent Workflow Memory (AWM)~\citep{wang2024agentworkflowmemory} distill experience into textual prompts. Continuous Memory (CoMEM)~\citep{wu2025comem, wu2025auto} compresses multimodal trajectories into continuous embeddings. Evaluations reveal the effectiveness of memory in single model setting.

\paragraph{RL-trained base models.}
We fine-tune models using the same data and hyperparameters as our method, but in a single-model training setup (``Base RL''). This serves as a controlled baseline to highlight the improvements introduced by our memory-augmented multi-agent design, particularly in scaling and generalization to unseen tasks.

\begin{table}[t]
\centering
\caption{Task success rates (\%) across WebVoyager, Mind2Web, and MMInA. Best open-source results in each domain are \textbf{bolded}. ``OOD'' denotes out-of-domain evaluation. $\Delta$ shows relative improvement over the backbone's single-model baseline.}
\label{tab:main_results}
\renewcommand{\arraystretch}{1.05}
\setlength{\tabcolsep}{1.5pt}
\scriptsize
\resizebox{\textwidth}{!}{%
\begin{tabular*}{\textwidth}{@{\extracolsep{\fill}} ll cccc ccccc cc @{}}
\toprule
\multicolumn{2}{c}{}
& \multicolumn{4}{c}{\textbf{WebVoyager (In Domain)}}
& \multicolumn{5}{c}{\textbf{Mind2Web \& MMInA (OOD)}}
\\
\cmidrule(lr){3-6} \cmidrule(lr){7-11}
\textbf{Backbone} & \textbf{Model/Method}
& \textbf{Amz} & \textbf{Cour} & \textbf{Recp} & \textbf{Map}
& \textbf{Info} & \textbf{Svc} & \textbf{Ent} & \textbf{Trav}
& \textbf{Wiki} & \textbf{Avg.} & $\Delta$ \\
\midrule

\multirow{3}{*}{Closed-Source}
& GPT-4o           & 24.4\% & 7.1\%  & 13.3\% & 36.6\% & 7.8\%  & 14.1\% & 3.4\%  & 19.4\% & 51.0\% & 19.7\% &\\
& Gemini-2.5-Pro   & 41.7\% & 28.6\% & 20.0\% & 53.7\% & 16.7\% & 22.4\% & 0.0\%  & 19.4\% & 50.0\% & 29.6\% &\\
& Claude-4         & 63.4\% & 28.6\% & 33.3\% & 70.0\% & 25.6\% & 40.0\% & 6.9\%  & 9.7\%  & 52.0\% & 36.6\% &\\
\midrule

\multirow{4}{*}{Open-Source}
& Qwen2.5-VL-32B   & 46.3\% & 26.2\% & 6.7\%  & 29.3\% & 14.1\% & 20.0\% & 6.9\%  & 9.7\%  & 43.0\% & 22.5\% &\\
& CogAgent          & 12.2\% & 9.5\%  & 26.7\% & 9.8\%  & 24.4\% & 8.2\%  & 13.8\% & 16.1\% & 21.0\% & 15.7\% &\\
& WebSight          & 24.4\% & 4.8\%  & 13.3\% & 29.3\% & 10.3\% & 3.5\%  & 3.4\%  & 0.0\%  & 12.0\% & 11.2\% &\\
& UI-TARS-1.5-7B   & 31.7\% & 16.7\% & 20.0\% & 31.7\% & 6.4\%  & 4.7\%  & 6.9\%  & 0.0\%  & 36.0\% & 17.1\% &\\
\midrule

\multirow{3}{*}{Qwen2.5-VL-3B}
& Baseline          & 7.3\%  & 11.9\% & 8.9\%  & 4.9\%  & 12.8\% & 4.7\%  & 6.9\%  & 6.5\%  & 7.0\%  & 7.9\%  & \improvement{7.9}{7.9}\\
& +Base RL         & \textbf{31.7\%} & \textbf{33.3\%} & 20.0\% & 29.3\% & 19.2\% & 16.5\% & 10.3\% & 6.5\% & \textbf{23.0\%} & 21.1\% & \improvement{21.1}{7.9}\\
\rowcolor{cyan!8}[2pt][2pt]
\cellcolor{white} & +Multi-Agent    & 22.0\% & 14.3\% & 4.4\%  & 14.6\% & 15.4\% & 5.9\%  & 13.8\% & 6.5\%  & 15.0\% & 12.4\% & \improvement{12.4}{7.9}\\
\rowcolor{cyan!8}[2pt][2pt]
\cellcolor{white} & +Multi-Agent RL & \textbf{31.7\%} & \textbf{33.3\%} & \textbf{24.4\%} & \textbf{41.5\%} & \textbf{21.8\%} & \textbf{17.6\%} & \textbf{13.8\%} & \textbf{12.9\%} & \textbf{23.0\%} & \textbf{24.0\%} & \improvement{24.0}{7.9}\\
\midrule

\multirow{7}{*}{Qwen2.5-VL-7B}
& Baseline          & 14.6\% & 2.4\%  & 15.9\% & 16.7\% & 9.0\%  & 11.8\% & 0.0\%  & 4.4\%  & 38.0\% & 12.5\% & \improvement{12.5}{12.5}\\
& +ReasoningBank   & 29.3\% & 9.5\%  & 6.7\%  & 29.3\% & 9.0\%  & 20.0\% & 3.4\%  & 6.5\%  & 44.0\% & 17.5\% & \improvement{17.5}{12.5}\\
& +AWM             & 17.1\% & 4.8\%  & 11.1\% & 29.3\% & 7.7\%  & 10.6\% & 0.0\%  & 6.5\%  & 31.0\% & 13.1\% & \improvement{13.1}{12.5}\\
& +CoMEM           & 24.4\% & 17.1\% & 8.9\%  & 34.1\% & 16.7\% & \textbf{23.5\%} & \textbf{10.3\%} & \textbf{12.9\%} & \textbf{47.0\%} & 21.7\% & \improvement{21.7}{12.5}\\
& +Base RL         & 46.3\% & 21.4\% & 24.4\% & 48.8\% & \textbf{23.1\%} & \textbf{23.5\%} & 6.9\%  & 6.5\%  & 34.0\% & 26.1\% & \improvement{26.1}{12.5}\\
\rowcolor{cyan!8}[2pt][2pt]
\cellcolor{white} & +Multi-Agent    & 56.1\% & 33.3\% & 24.4\% & 48.8\% & 15.4\% & 21.2\% & 3.4\%  & 3.2\%  & 41.0\% & 27.4\% & \improvement{27.4}{12.5}\\
\rowcolor{cyan!8}[2pt][2pt]
\cellcolor{white} & +Multi-Agent RL & \textbf{68.3\%} & \textbf{38.1\%} & \textbf{33.3\%} & \textbf{63.4\%} & 21.8\% & \textbf{23.5\%} & \textbf{10.3\%} & \textbf{12.9\%} & 44.0\% & \textbf{35.1\%} & \improvement{35.1}{12.5}\\
\midrule

\multirow{7}{*}{Qwen3-VL-8B}
& Baseline          & 36.6\% & 16.7\% & 13.3\% & 43.9\% & 9.0\%  & 20.0\% & 10.3\% & 9.7\%  & 38.0\% & 21.9\% & \improvement{21.9}{21.9}\\
& +ReasoningBank   & 36.6\% & 11.9\% & 17.8\% & 31.7\% & 10.3\% & 24.7\% & 17.2\% & 6.5\%  & 48.0\% & 22.7\% & \improvement{22.7}{21.9}\\
& +AWM             & 39.0\% & 19.0\% & 8.9\%  & 31.7\% & 12.5\% & 16.5\% & 6.9\%  & 16.1\% & 48.0\% & 22.1\% & \improvement{22.1}{21.9}\\
& +CoMEM           & 43.9\% & 19.0\% & 17.8\% & 34.1\% & 12.5\% & 17.9\% & 3.4\%  & \textbf{19.4\%} & 42.0\% & 23.3\% & \improvement{23.3}{21.9}\\
& +Base RL         & 46.3\% & 38.1\% & 33.3\% & 48.8\% & 25.6\% & 18.8\% & 10.3\% & 6.5\%  & 45.0\% & 30.3\% & \improvement{30.3}{21.9}\\
\rowcolor{cyan!8}[2pt][2pt]
\cellcolor{white} & +Multi-Agent    & 56.1\% & 33.3\% & 28.9\% & \textbf{58.5\%} & 21.8\% & \textbf{27.1\%} & \textbf{20.7\%} & 16.1\% & \textbf{54.0\%} & 35.2\% & \improvement{35.2}{21.9}\\
\rowcolor{cyan!8}[2pt][2pt]
\cellcolor{white} & +Multi-Agent RL & \textbf{63.4\%} & \textbf{42.9\%} & \textbf{37.8\%} & 53.7\% & \textbf{26.9\%} & 23.5\% & 17.2\% & 9.7\%  & 51.0\% & \textbf{35.8\%} & \improvement{35.8}{21.9}\\

\bottomrule
\end{tabular*}
}
\end{table}

\subsection{Main results}
\label{sec:main_results}

Table~\ref{tab:main_results} presents comprehensive evaluation results. We highlight three key findings.

\paragraph{Multi-agent architecture enables strong performance.}
The training-free Qwen2.5-VL-7B + Multi-Agent achieves 27.4\% overall, more than doubling its single-model baseline (12.5\%) and surpassing GPT-4o (19.7\%) and specialized fine-tuned agents like UI-TARS-1.5-7B (17.1\%). This validates that shifting cognitive load from a monolithic model to a collaborative system releases latent reasoning capabilities. Note that although our multi-agent setup uses more inference compute (three model calls per step); the gains substantially exceed what would be expected from simply scaling inference compute in a single model. More discussions about inference compute analysis can be found in Appendix \ref{append:inference_compute}.

\paragraph{Planner-centric RL further improves performance.}
Applying RL to the multi-agent setup yields consistent improvements across all model scales. For Qwen2.5-VL-7B, RL boosts success from 27.4\% to 35.1\%, a 28\% relative improvement. For Qwen2.5-VL-3B, RL nearly doubles the success rate from 12.4\% to 24.0\%, bridging the gap between reasoning and execution in capacity-constrained regimes.

\paragraph{Robust generalization to out-of-domain tasks.}
Our method generalizes to out-of-domain benchmarks, confirming that we are not overfitting to training domains. On MMInA (Wiki), our Qwen3-VL-8B agent achieves 51.0\%, outperforming memory-augmented baselines. The Planner's learned reasoning ability transfers across unseen interface layouts.

\begin{table*}[t]
\centering
\caption{Performance on OSWorld across different methods.}
\label{tab:osworld_performance}

\begingroup
\small
\setlength{\tabcolsep}{2.7pt}
\renewcommand{\arraystretch}{1.05}

\begin{adjustbox}{max width=\textwidth,center}
\begin{tabular}{@{}lcccccccccccccc@{}}
\toprule
& \multicolumn{4}{c}{\textbf{Office}} & \multicolumn{4}{c}{\textbf{Daily}} & \multicolumn{3}{c}{\textbf{Professional}} & \multicolumn{2}{c}{\textbf{Others}} & \\
\cmidrule(lr){2-5} \cmidrule(lr){6-9} \cmidrule(lr){10-12} \cmidrule(lr){13-14}
\textbf{Method} & \textbf{Writer} & \textbf{Calc} & \textbf{Impress} & \textbf{Avg.} & \textbf{Thunder.} & \textbf{VLC} & \textbf{Chrome} & \textbf{Avg.} & \textbf{GIMP} & \textbf{VSCode} & \textbf{Avg.} & \textbf{M-Apps} & \textbf{OS} & \textbf{Overall} \\
\midrule
Qwen2.5-VL-7B & 4.3\% & 2.1\% & 2.1\% & 2.6\% & 0.0\% & \textbf{5.9\%} & \textbf{11.1\%} & 7.8\% & 0.0\% & 0.0\% & 0.0\% & 2.0\% & 12.5\% & 3.9\% \\
+ReasoningBank & 8.7\% & 2.1\% & 2.1\% & 3.4\% & 13.3\% & 0.0\% & 6.5\% & 6.4\% & 11.5\% & \textbf{4.6\%} & \textbf{8.3\%} & 0.9\% & 12.5\% & 4.7\% \\
\rowcolor{orange!15}
+Multi-Agent & \textbf{8.7\%} & 0.0\% & \textbf{4.3\%} & 3.4\% & 13.3\% & 0.0\% & 8.7\% & \textbf{7.9\%} & 11.5\% & 0.0\% & 6.3\% & \textbf{3.1\%} & 12.5\% & \textbf{5.3\%} \\
\midrule
Qwen3-VL-8B & 27.3\% & 4.3\% & \textbf{19.0\%} & \textbf{14.6\%} & 40.0\% & 17.7\% & 28.3\% & 28.2\% & 30.8\% & 18.2\% & 25.0\% & 7.4\% & 29.2\% & 18.1\% \\
+ReasoningBank & 18.2\% & 8.5\% & 10.6\% & 11.2\% & \textbf{53.3\%} & 23.5\% & 26.1\% & 30.8\% & 36.0\% & \textbf{27.3\%} & 31.9\% & 6.3\% & 41.7\% & 19.0\% \\
\rowcolor{orange!15}
+Multi-Agent & 27.3\% & \textbf{8.7\%} & 10.4\% & 13.0\% & 46.7\% & \textbf{29.4\%} & \textbf{34.8\%} & \textbf{35.9\%} & \textbf{46.2\%} & 22.7\% & \textbf{35.4\%} & \textbf{9.6\%} & 41.7\% & \textbf{22.1\%} \\
\bottomrule
\end{tabular}
\end{adjustbox}

\endgroup
\end{table*}

\begin{table*}[t]
\centering
\caption{Performance comparison on MCPBench across different methods.}
\label{tab:mcpbench_result}

\begingroup
\small
\setlength{\tabcolsep}{3pt}
\renewcommand{\arraystretch}{1.2}

\begin{adjustbox}{max width=\textwidth,center}
\begin{tabular}{@{}lcccccccccc@{}}
\toprule
& \multicolumn{2}{c}{\textbf{Task Completion}} & \multicolumn{2}{c}{\textbf{Tool Usage}} & \multicolumn{2}{c}{\textbf{Planning Eff.}} & \multicolumn{4}{c}{\textbf{Execution Fidelity}} \\
\cmidrule(lr){2-3} \cmidrule(lr){4-5} \cmidrule(lr){6-7} \cmidrule(lr){8-11}
\textbf{Model} &
\textbf{Task} & \textbf{Info} &
\textbf{Tool} & \textbf{Param} &
\textbf{Dep} & \textbf{Parallel} &
\textbf{Schema} & \textbf{Valid Tool} & \textbf{Exec} & \textbf{Fail} \\
&
\textbf{Fulfill.} & \textbf{Ground.} &
\textbf{Appr.} & \textbf{Acc.} &
\textbf{Aware.} & \textbf{Eff.} &
\textbf{Compli.} & \textbf{Rate} & \textbf{Success} & \textbf{Rate} \\
\midrule
Qwen2.5-VL-7B & 4.72 & 4.54 & 4.46 & 4.42 & 4.08 & 3.73 & 0.57 & 0.57 & 0.52 & 0.05 \\
+ReasoningBank & 6.09 & 5.84 & 5.50 & 5.67 & 5.71 & 4.70 & 0.81 & 0.74 & 0.68 & 0.10 \\
\rowcolor{orange!15}
+Planner+Actor & 6.78 & 6.68 & \textbf{6.70} & \textbf{6.51} & \textbf{5.83} & \textbf{5.26} & \textbf{0.89} & \textbf{0.92} & \textbf{0.91} & \textbf{0.01} \\
\rowcolor{orange!15}
+Planner+Actor+Memory & \textbf{6.82} & \textbf{6.69} & 5.86 & 6.20 & 5.23 & 4.85 & 0.80 & 0.74 & 0.70 & 0.04 \\
\midrule
Qwen3-VL-8B & 5.67 & 5.68 & 5.69 & 5.63 & 5.69 & 5.42 & 0.72 & 0.71 & 0.70 & 0.01 \\
+ReasoningBank & 6.93 & 7.04 & 6.96 & 6.84 & 6.79 & 6.57 & 0.90 & 0.85 & 0.81 & 0.05 \\
\rowcolor{orange!15}
+Planner+Actor & \textbf{8.71} & \textbf{8.59} & \textbf{8.57} & \textbf{8.60} & 8.22 & \textbf{8.09} & \textbf{1.00} & \textbf{1.00} & \textbf{1.00} & \textbf{0.00} \\
\rowcolor{orange!15}
+Planner+Actor+Memory & 8.10 & 8.26 & 8.29 & 8.15 & \textbf{8.26} & 7.51 & \textbf{1.00} & \textbf{1.00} & \textbf{1.00} & \textbf{0.00} \\
\bottomrule
\end{tabular}
\end{adjustbox}

\endgroup
\end{table*}

\subsection{Generalization to OS and tool-use tasks}

We evaluate on OSWorld~\citep{xie2024osworldbenchmarkingmultimodalagents} (OS-level desktop tasks, Table~\ref{tab:osworld_performance}) and MCPBench~\citep{wang2025mcpbench} (multi-step tool-use tasks, Table~\ref{tab:mcpbench_result}). Our multi-agent framework consistently outperforms baselines on both benchmarks across Qwen2.5-VL-7B and Qwen3-VL-8B backbones. Notably, with Qwen3-VL-8B on MCPBench, our approach achieves perfect Execution Fidelity scores (1.00), demonstrating that the planner-centric architecture generalizes well beyond web navigation.


\subsection{Further analysis}

\begin{wrapfigure}{r}{0.4\textwidth}
    \centering
    \includegraphics[width=\linewidth]{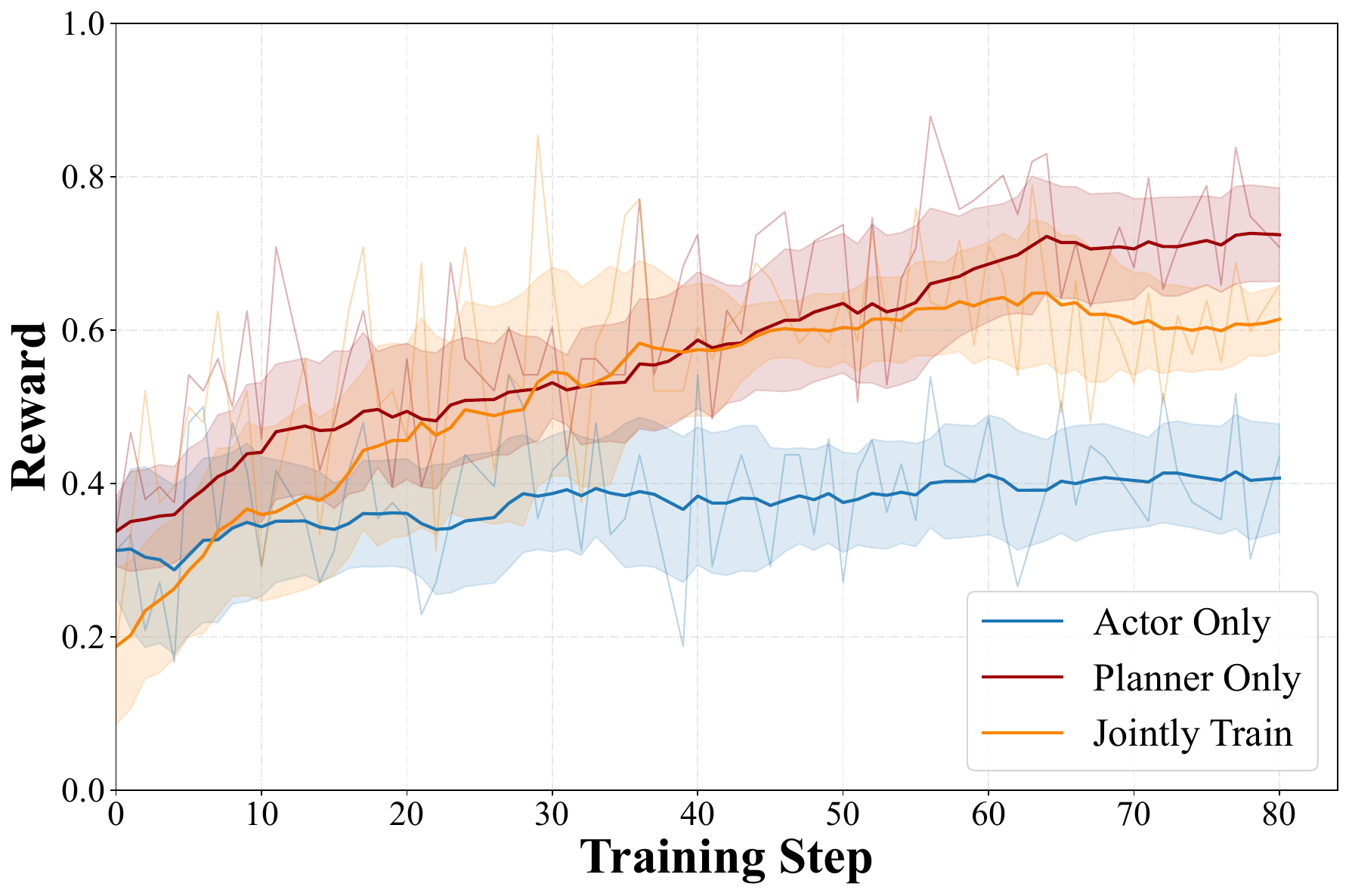}
    \caption{Reward curves under different RL training strategies.}
    \label{fig:training_comparison}
\end{wrapfigure}

\paragraph{RL training strategy ablation.}
We ablate which components are optimized during training: (1) Planner-only (ours), (2) Actor-only with frozen Planner, and (3) joint training of Planner and Actor with shared parameters.
As shown in Table~\ref{tab:train_strategy} and Figure~\ref{fig:training_comparison}, training the Planner alone achieves the best overall performance and most stable reward improvement.
\textit{Joint training with shared parameters introduces competing objectives}: the model must both reason over long-horizon plans and execute precise low-level actions, leading to role confusion. We note that an alternative joint-training approach with separate parameters and alternating updates may alleviate this issue, but would approximately double training cost.
\textit{Actor-only training cannot improve reasoning quality}: the Actor operates under the Planner's guidance and has limited ability to improve plan correctness.
\textit{Planner-only training allows focused improvement of high-level reasoning} while the frozen Actor reliably follows instructions, yielding the best division of labor.

\begin{table}[h]
\centering
\caption{Task success rates with different training strategies.}
\label{tab:train_strategy}
\renewcommand{\arraystretch}{1}
\setlength{\tabcolsep}{6pt}
{\small
\begin{tabular}{lcccc}
\toprule
\textbf{Training Strategy} & \textbf{Amazon} & \textbf{Map} & \textbf{Cour} & \textbf{Recp} \\
\midrule
Planner-only (Ours) & \textbf{68.3\%} & \textbf{58.5\%} & \textbf{38.1\%} & \textbf{28.9\%} \\
\hline
Actor-only & 48.8\% & 58.5\% & 31.0\% & 22.2\% \\
Joint Training & 48.8\% & 58.5\% & 42.9\% & 13.3\% \\
\bottomrule
\end{tabular}
}
\end{table}

\paragraph{Ablation on continuous memory.}
We compare our method with and without continuous memory (CoMEM). In Table~\ref{tab:comem_ablation}, CoMEM consistently improves success rates, with +7.1\%, +4.4\%, and +4.9\% gains on Coursera, Allrecipes, and Google Maps, respectively, confirming that continuous multimodal memory complements discrete text summaries.

\begin{table}[t]
\centering
\caption{Performance with and without continuous memory across different domains.}
\label{tab:comem_ablation}
\renewcommand{\arraystretch}{1}
\setlength{\tabcolsep}{6pt}
{\small
\begin{tabular}{lcccc}
\toprule
\textbf{Setting} & \textbf{Map} & \textbf{Cour} & \textbf{Recp} & \textbf{Overall}\\
\midrule
w/ CoMEM & \textbf{63.4\%} & \textbf{45.2\%} & \textbf{33.3\%} & \textbf{47.3\%} \\
w/o CoMEM & 58.5\% & 38.1\% & 28.9\% & 36.9\% \\
\bottomrule
\end{tabular}
}
\end{table}

\section{Related work}

\paragraph{Multi-agent systems for digital tasks.}


Advances in vision–language models (VLMs) have enabled agents to ground language instructions in visual interfaces with increasing accuracy~\citep{hong2024cogagentvisuallanguagemodel}. Single-agent systems tackle end-to-end web and desktop tasks by directly interacting with live interfaces~\citep{he2024webvoyager, openai2025computerusingagent}. Tool-use agents further integrate reasoning with tool invocation, supported by frameworks like the Model Context Protocol (MCP)~\citep{yao2023reactsynergizingreasoningacting, schick2023toolformerlanguagemodelsteach, anthropic2024mcp}.  
Multi-agent systems enhance task performance by distributing roles among specialized model instances~\citep{wu2023autogenenablingnextgenllm, qian2024chatdevcommunicativeagentssoftware}. Plan-and-Act~\citep{erdogan2025planandactimprovingplanningagents} decomposes tasks into planner and actor roles, showing planning as the main contributor to performance. Frameworks like CES~\citep{ces2025scheduler} and AgentFlow~\citep{li2025agentflow} focus on high-level module training and planner-centric optimization for tool-use scenarios, while COLA~\citep{zhao2025colascalablemultiagentframework} explores multi-agent setups for Windows UI automation. 

\paragraph{Long-horizon planning for agents.}

Long-horizon tasks often rely on reasoning–acting loops, such as ReAct~\citep{yao2023reactsynergizingreasoningacting}, or plan-then-execute frameworks~\citep{erdogan2025planandactimprovingplanningagents}, where reasoning is interleaved with action execution. Methods like memory augmentation~\citep{wang2024agentworkflowmemory, packer2024memgptllmsoperatingsystems}, test-time scaling~\citep{yao2023treethoughtsdeliberateproblem, wang2023selfconsistencyimproveschainthought}, and search-guided backtracking~\citep{zhou2024languageagenttreesearch, putta2024agentqadvancedreasoning} enhance performance in sparse-reward environments. Reflective memory approaches~\citep{shinn2023reflexionlanguageagentsverbal, wu2025backtrackagentenhancingguiagent} and decomposition-based replanning~\citep{prasad2024adaptasneededdecompositionplanning} further improve recovery in complex tasks. 

\paragraph{Training for agents.}
Supervised fine-tuning (SFT) learns reliable action priors~\citep{sun2025osgenesisautomatingguiagent, trabucco2025instainternetscaletrainingagents}, but expert trajectories are scarce. Automatic pipelines explore environments and generate training data~\citep{pahuja2025explorerscalingexplorationdrivenweb, sun2025osgenesisautomatingguiagent}, though quality remains an open issue.
RL from online interaction has shown promise for long-horizon interactive tasks~\citep{wei2025webagentr1trainingwebagents, qi2025webrltrainingllmweb, li2025agentflow}, though sparse rewards and long trajectories make training challenging. Process reward models provide denser step-level signals~\citep{chae2025webshepherdadvancingprmsreinforcing}. 

\section{Conclusion}
We studied the challenges of long-horizon task completion for autonomous agents and identified high-level planning as the primary bottleneck through systematic compute-allocation analysis. By deploying a multi-agent framework that separates planning, memory management, and action execution, we achieved substantial improvements across web, OS, and tool-use benchmarks. Our scaling analysis quantitatively demonstrates that the planner is the dominant component: its scaling coefficient matches that of scaling all modules simultaneously. Building on this insight, our planner-centric RL approach yields robust and compute-efficient gains. We hope this work provides practical guidance for compute allocation in multi-agent systems and inspires further research on planner-first designs under constrained budgets.

\bibliography{colm2026_conference}
\bibliographystyle{colm2026_conference}

\newpage
\appendix
\section{Appendix}

\subsection{Ethics statement}
This work aims to improve the reasoning and memory capabilities of autonomous agents in a planner-centric multi-agent framework. All web-browser experiments are conducted on public websites (\eg Amazon, Google Maps, wikipedia, etc.), commonly used in prior research. To ensure safety, we employ a VLM to detect when the agent navigates to login or sensitive pages; in such cases, the agent is redirected to the main page. No login-protected content or private data are involved. For Operation system tasks, we use environment in virtual machine provided by OSWorld~\citep{xie2024osworldbenchmarkingmultimodalagents}. For tool-use MCP tasks, all tools are officially provided my different websites.

While our study poses no immediate privacy or ethical risks, we recognize that deploying web agents at scale requires strong safeguards. Without proper constraints, agents may unintentionally perform harmful or unauthorized actions. Future real-world applications should incorporate safety mechanisms and human oversight to ensure responsible and ethical use.

\subsection{Reproducibility statement}
Training hyperparameters are in Appendix~\ref{append:training_params}; task generation and filtering in Appendix~\ref{append:task_generation}; VLM-as-judge prompt in Appendix~\ref{append:judge_prompt}; all agent prompts in Appendix~\ref{append:prompts}. Evaluation uses standard protocols on publicly available benchmarks. Our code will be publicly released to support the reproducibility.

\subsection{Action space definition}
\label{append:action_space}

The action space $\mathcal{A}$ consists of parameterized interaction primitives. Each action is a tuple $(\texttt{type}, \texttt{params})$ where $\texttt{type}$ specifies the action type, and $\texttt{params}$ specifies the target and additional arguments. The action space is divided into two categories based on the nature of the task: actions for GUI-based environments (web and OS) and actions for tool-use tasks.

\paragraph{Actions for GUI tasks.}  
GUI-based actions involve interacting with visual user interfaces for both web and OS environments. These actions are executed via Playwright\footnote{\url{https://playwright.dev/}} and PyAutoGUI\footnote{\url{https://pyautogui.readthedocs.io/}} for OS operation. Concretely, the following action types are available:
\begin{itemize}[nosep, leftmargin=*]
    \item $\texttt{Click}(\texttt{element\_id})$: Click on the element identified by its index in the accessibility tree.
    \item $\texttt{Type}(\texttt{element\_id}, \texttt{text})$: Enter text into the specified input field.
    \item $\texttt{Scroll}(\texttt{direction}, \texttt{amount})$: Scroll the page in the given direction.
    \item $\texttt{Select}(\texttt{element\_id}, \texttt{option})$: Choose an option from a dropdown menu.
    \item $\texttt{Stop}(\texttt{answer})$: Terminate the episode and return a final answer.
\end{itemize}

\paragraph{Actions for tool-use tasks.}  
Tool-use actions focus on interacting with external tools through API-based commands or invoking operations defined by the tool. These actions are designed for multi-tool integration systems such as MCP (Model Context Protocol)~\citep{wang2025mcpbench}. Tool-use actions include:
\begin{itemize}[nosep, leftmargin=*]
    \item $\texttt{ToolInvoke}(\texttt{tool\_name}, \texttt{tool\_params})$: Invoke a specific tool with the provided parameters. The argument $\texttt{tool\_name}$ specifies which tool to use, and $\texttt{tool\_params}$ supplies the arguments required by the tool (\eg API endpoint, input data).
    \item $\texttt{Stop}(\texttt{answer})$: Terminate the episode and return a final answer.
\end{itemize}
Tool-use actions are executed through the MCP framework, which standardizes interactions by enabling seamless tool calls and parameter passing across diverse tool APIs.

\subsection{Experiment details}
\label{append:experiment_details}

\subsubsection{Training task generation}
\label{append:task_generation}

We develop a pipeline for automated task generation and filtering using Qwen2.5-VL-32B. For each webpage screenshot, we generate contextual descriptions and propose $K=10$ structured tasks with difficulty labels. Tasks are filtered through: (1) automatic executability checks with $N=6$ rollouts per task, retaining only those with non-zero success rate; and (2) manual quality assurance. This yields 320 high-quality tasks spanning multiple domains and difficulty levels.

\subsubsection{Training parameters}
\label{append:training_params}
We list our hyperparameter setting for planner-centric RL training in table \ref{tab:rl_params}. All our ablation RL training use the identical parameter setting.

\begin{table}[h]
\centering
\caption{Training hyperparameters for planner-centric RL.}
\label{tab:rl_params}
\begin{tabular}{ll}
\toprule
\textbf{Parameter} & \textbf{Value} \\
\midrule
Planner Temperature & 0.5 \\
Learning Rate & $2 \times 10^{-6}$ \\
Batch Size & 6 \\
Rollouts per Sample & 8 \\
KL-Divergence Coefficient ($\beta$) & 0.1 \\
\bottomrule
\end{tabular}
\end{table}

\subsubsection{Memory construction details}
\label{append:memory_construct}

\paragraph{Trajectories Collection.} We leverage the autonomous task generation and rollout framework described by~\citet{wu2025auto} to collect trajectories across websites including Amazon, Google Maps, Coursera, and Allrecipes. Following the same setup as in Section~\ref{append:task_generation}, we perform $N = 6$ rollouts per task and retain only successful trajectories.

\paragraph{Discrete memory.} We use Qwen2.5-VL-7B to extract key steps from successful trajectories, identifying important subgoals, salient UI elements, and critical action summaries. This yields compact symbolic representations provided to the Planner as context.

\paragraph{Continuous memory.}
Following~\citet{wu2025comem}, we use a VLM-based encoder consisting of the trained Planner model and a pretrained Q-Former. The trajectory is processed through the Planner to obtain hidden states, which are compressed into $n=8$ latent tokens:
$M_c = \text{QFormer}(\text{Planner}(h_{-1})) \in \mathbb{R}^{8 \times d}$.
These tokens are prepended to the Planner's input embeddings during inference.

\subsection{Baseline setup and comparison context}
\label{append:baseline_setup}

\paragraph{Closed-source baseline configuration.}
All closed-source models (GPT-4o, Gemini-2.5-Pro, Claude-4) are evaluated in a \emph{single-model, screenshot-only} setup. Specifically:
\begin{itemize}[nosep, leftmargin=*]
    \item \textbf{Observation}: The model receives a high-resolution screenshot of the current webpage at each step, along with the accessibility tree (element IDs and text content). No DOM HTML is provided.
    \item \textbf{Prompting}: We use ReAct-style prompting where the model is asked to reason about the current state and then produce a single action.
    \item \textbf{No external tools}: The models do not have access to any external tools, APIs, or agentic scaffolding beyond the basic action space (Click, Type, Scroll, Select, Stop).
    \item \textbf{Step limit}: All models are limited to a maximum of 15 steps per task.
\end{itemize}

\paragraph{Context on absolute performance levels.}
We note that our reported absolute numbers on benchmarks like WebVoyager are lower than some recently reported state-of-the-art results. For example, systems such as OpenAI Operator~\citep{openai2025operatorsystemcard} and browser-use frameworks with extended agentic scaffolding have reported substantially higher success rates on WebVoyager. This discrepancy is due to several key differences in our evaluation setting:
\begin{enumerate}[nosep, leftmargin=*]
    \item \textbf{Dynamic web environments}: Real-world websites change continuously, such as page layouts, product listings, UI elements, and even entire website structures are updated frequently. Tasks that were valid months ago may no longer be feasible in their original form. Our evaluation is conducted on live websites at the time of our experiments, meaning the difficulty may differ from prior evaluations conducted on older snapshots.
    \item \textbf{Step limit}: We enforce a strict 15-step limit for all methods and models, which is more restrictive than some prior work that allows 50-100 steps.
    \item \textbf{No website-specific optimization}: We do not fine-tune prompts, action spaces, or memory banks to specific websites. Our framework uses the same configuration across all evaluation domains.
    \item \textbf{Fair comparison environment}: All methods and models in our evaluation, including closed-source baselines, open-source models, memory-augmented variants, and our multi-agent framework, are evaluated under identical conditions (same websites, same tasks, same step limits, same evaluation time window). This ensures that performance differences reflect genuine method improvements rather than evaluation setup advantages.
\end{enumerate}
The focus of our work is not to achieve the highest absolute numbers, but rather to provide a controlled and fair analysis of how compute allocation and training strategy affect multi-agent system performance. Our key findings: the planner is the dominant bottleneck and that planner-centric RL is effective, hold regardless of absolute performance levels.

\subsection{VLM-as-judge evaluation details}
\label{append:judge_prompt}

\paragraph{Overview.}
Using LLMs and VLMs as automated evaluators has become a widely adopted paradigm across NLP and agent research. \citet{zheng2023judging} introduced the LLM-as-a-judge framework and demonstrated that strong LLMs such as GPT-4 can achieve over 80\% agreement with human annotators, matching inter-annotator agreement levels. Since then, this approach has been extensively adopted in agent evaluation: WebVoyager~\citep{he2024webvoyager} employs GPT-4V to judge task completion from screenshots, replacing fragile rule-based heuristics; VisualWebArena~\citep{koh2024visualwebarenaevaluatingmultimodalagents} uses multimodal LLMs to assess open-ended visual web tasks; \citet{pan2024autonomousevalagent} propose using LLMs for both evaluating and refining agent trajectories; and several RL-based agent training methods~\citep{qi2025webrltrainingllmweb, liu2025flowgrpotrainingflowmatching} rely on LLM/VLM judges to provide trajectory-level reward signals. A recent comprehensive survey~\citep{gu2024surveyllmasjudge} systematically reviews the reliability, biases, and mitigation strategies of LLM-as-a-judge systems, confirming that with appropriate prompting and aggregation, VLM-based evaluation provides a scalable and reliable alternative to human annotation for agent tasks.

Following this established practice, our VLM evaluator (Qwen2.5-VL-32B) receives the task description, the full sequence of screenshots, all planner outputs, and the final answer. It evaluates on three criteria: (1) \textbf{Task correctness}: whether the final state satisfies the original task requirements; (2) \textbf{Reasoning coherence}: whether the planning steps form a logical sequence toward the goal; (3) \textbf{Interaction efficiency}: whether the agent avoids unnecessary repetition or wasted actions. The evaluator assigns a holistic score of $r \in \{1, 3, 5\}$ where 1 = failure, 3 = partial success, 5 = full success. We use $K=3$ independent evaluations and take the mode to reduce variance. Evaluation prompt can be found in Appendix \ref{append:prompts}.

\paragraph{Human--VLM agreement analysis.}

Prior work has shown that LLM-based judges can achieve human-level agreement when properly calibrated~\citep{zheng2023judging, gou2025mind2web}, but domain-specific validation remains important, particularly for visually grounded agent tasks where evaluation requires jointly reasoning over screenshots, action sequences, and task semantics. To validate the reliability of our VLM-as-judge reward in this specific setting, we conduct a human agreement study: three human annotators independently score a random subset of trajectories using the same rubric as the VLM evaluator (1 = failure, 3 = partial, 5 = success). We then compute agreement between the majority human label and the VLM judge's mode score across $K=3$ evaluations. Results are shown in Table~\ref{tab:judge_agreement}.

\begin{table}[h]
\centering
\caption{Agreement between VLM-as-judge (Qwen2.5-VL-32B) and human annotations on a sample of evaluation trajectories.}
\label{tab:judge_agreement}
\begin{tabular}{lcc}
\toprule
\textbf{Metric} & \textbf{Value} \\
\midrule
Number of trajectories evaluated & 100 \\
Exact agreement rate & 88.3\% \\
Agreement within $\pm 1$ level & 96.7\% \\
\bottomrule
\end{tabular}
\end{table}

\subsection{Inference compute analysis}
\label{append:inference_compute}

\begin{table}[h]
\centering
\caption{Inference compute comparison between single-model and multi-agent setups.}
\label{tab:efficiency_analysis}
\resizebox{\textwidth}{!}{%
\begin{tabular}{lcccccccc}
\toprule
\textbf{Backbone} & \multicolumn{2}{c}{\textbf{Amazon}} & \multicolumn{2}{c}{\textbf{Google Map}} & \multicolumn{2}{c}{\textbf{Coursera}} & \multicolumn{2}{c}{\textbf{Allrecipes}} \\
                  & Time (s) & Acc.    & Time (s) & Acc.    & Time (s) & Acc.    & Time (s) & Acc. \\
\midrule
GPT-4o            & 323.4    & 24.4\%  & 248.6    & 36.6\%  & 272.2    & 7.1\%   & 295.4    & 13.3\% \\
Claude-4          & 100.7    & 63.4\%  & 91.6     & 70.0\%  & 161.1    & 28.6\%  & 174.1    & 33.3\% \\
Qwen2.5-VL-7B     & 163.8    & 14.6\%  & 162.8    & 16.7\%  & 134.4    & 2.4\%   & 209.1    & 15.9\% \\
+ Multi-Agent    & \textbf{154.3}    & \textbf{68.3\%}  & \textbf{162.6}    & \textbf{63.4\%}  & \textbf{153.7}    & \textbf{38.1\%}  & \textbf{220.5}    & \textbf{33.3\%} \\
\bottomrule
\end{tabular}}
\end{table}

Here we analyze the average time cost in seconds for each task across different domains, and average accuracy. The results in Table \ref{tab:efficiency_analysis} demonstrate that the multi-agent framework achieves comparable or even reduced inference time compared to single-model baselines, despite additional computation from the Planner, Actor, and Memory Manager. On domains such as Coursera and Amazon, the multi-agent system significantly reduces time costs (\eg 163.8s to 153.7s on Coursera) by optimizing decision paths, making them more focused and efficient.

Moreover, the multi-agent setup consistently improves accuracy across all domains. For example, on Google Maps, accuracy rises from 16.7\% (Qwen2.5-VL-7B) to 63.4\%, while maintaining similar inference time. This shows that the multi-agent approach better leverages memory and collaboration to reduce unnecessary steps and enhance task success rates, making it both an efficient and reliable choice for real-world applications.

\subsection{Evaluation benchmarks}
\label{append:evaluation_benchmark}

\textbf{WebVoyager}~\citep{he2024webvoyager} includes real-world tasks from 15 websites.
\textbf{Mind2Web}~\citep{deng2023mind2webgeneralistagentweb} features over 2,000 tasks from 137 websites across 31 domains; we use the first 100 tasks from test-domain and test-website splits.
\textbf{MMInA}~\citep{tian2025mminabenchmarkingmultihopmultimodal} assesses agents on real websites across information searching, shopping, and travel; we evaluate on the Wikipedia domain using 100 test tasks.
\textbf{OSWorld}~\citep{xie2024osworldbenchmarkingmultimodalagents} requires agents to complete tasks in desktop applications via multi-turn interaction.
\textbf{MCPBench}~\citep{wang2025mcpbench} tests multi-step tasks requiring tool use via MCP servers.
For WebVoyager and Mind2Web, we use Qwen2.5-VL-32B as the LLM-as-judge. For MMInA, we compare the agent's final answer with ground truth using an LLM.

\subsection{Limitations}
While our study provides significant insights into employing multi-agent frameworks for diverse domain-specific tasks, there are several limitations to note. In the joint training ablation, the Actor and Planner share the same parameters, and we did not validate the optimization of two distinct models simultaneously. This decision was made due to computational resource constraints. Additionally, the experiments were restricted to a maximum of 15 steps for task execution, a setting imposed by cost and time limitations. Expanding the step limit in future work could provide a more comprehensive evaluation of the framework's effectiveness in longer tasks and complex scenarios.

\subsection{Case studies}
\label{sec:case_study}

\subsubsection{Case study for training-free planner}
\label{append:case_study_planner}

\begin{figure}[h]
    \centering
    \includegraphics[width=0.6\linewidth]{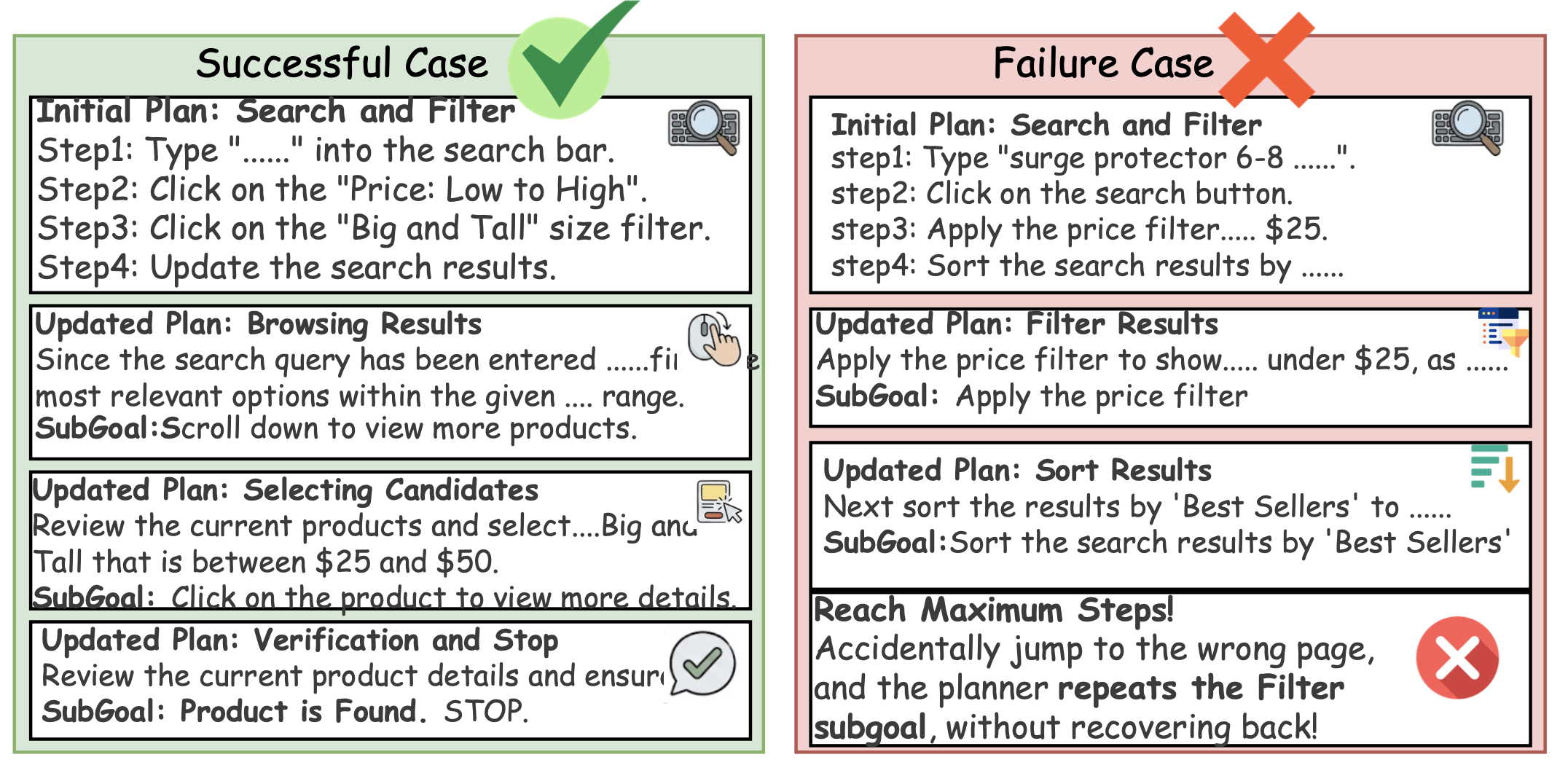}
    \caption{Success and failure cases of the training-free planner.}
    \label{fig:planner_case}
\end{figure}

We analyze a representative task: finding a surge protector on Amazon with specific constraints. In the successful case, the Planner effectively decomposes the task into subgoals and adapts based on feedback. In the failure case, the actor clicks into a wrong page and the Planner cannot recover, getting stuck repeating the same subgoal. This highlights the need for RL to improve adaptive decision-making.

\subsubsection{Case study for multi-agent vs.\ single-agent}
\label{append:case_study_multi-agent}

\begin{figure*}[t]
    \centering
    \includegraphics[width=0.8\textwidth]{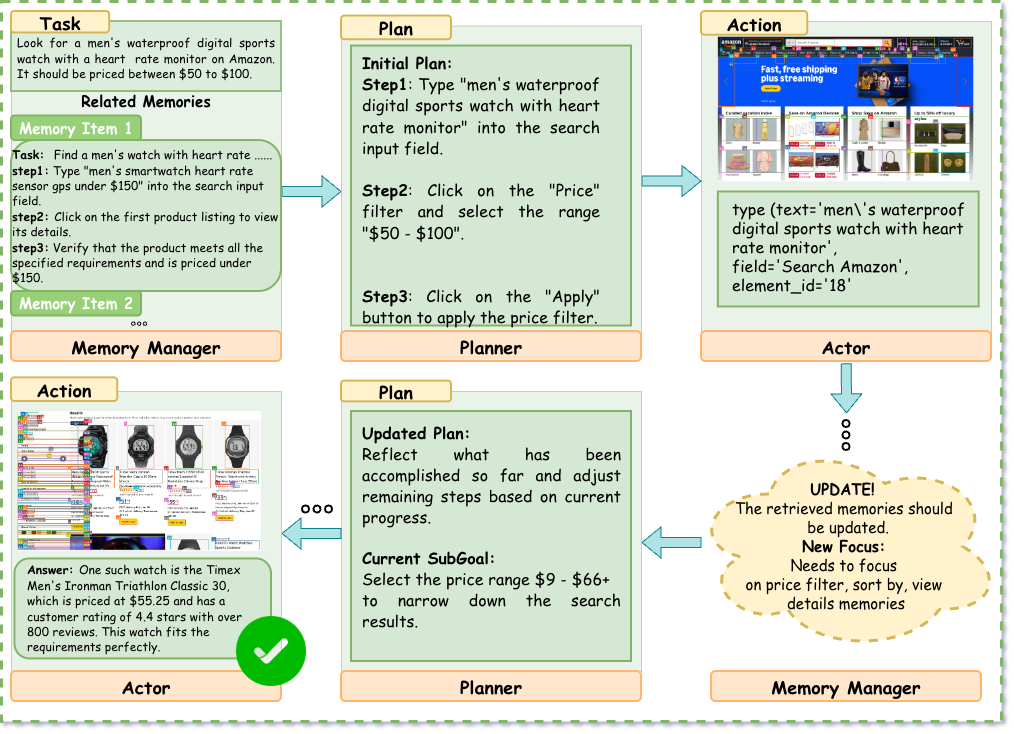}
    \includegraphics[width=0.8\textwidth]{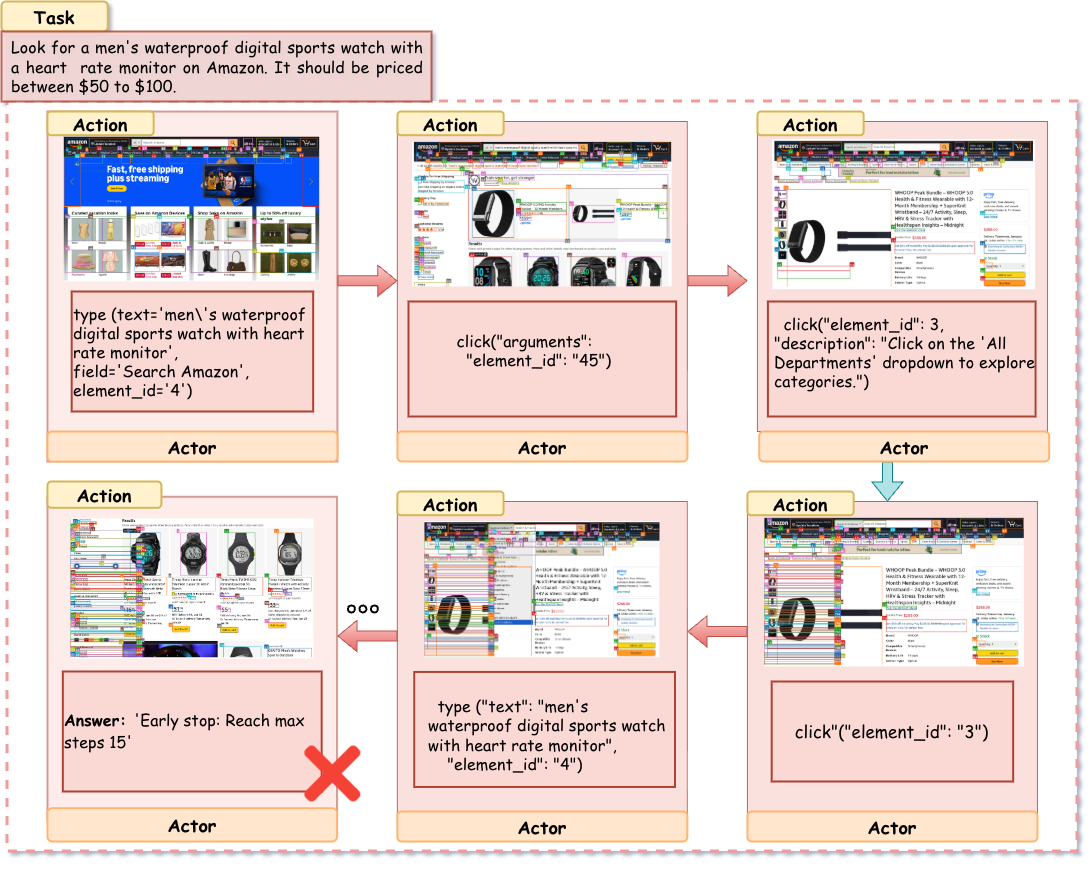}
    \caption{\textbf{Qualitative comparison on Amazon.} Top: The multi-agent framework successfully decomposes the task with the Planner instructing the Actor to apply price filters. Bottom: The single-agent baseline fails to utilize UI filters, repeating search queries until reaching the step limit.}
    \label{fig:case_study}
\end{figure*}

The single-agent baseline relies solely on search queries, failing to leverage sidebar filters and entering repetitive loops. Our multi-agent framework demonstrates structured reasoning: the Planner decomposes the query, dynamically adjusts the plan after observing mixed results, and instructs the Actor to apply specific constraints, leading to successful retrieval in fewer steps.

\clearpage

\subsection{Instruction prompts}
\label{append:prompts}

\begin{PromptBox}{Plan generation and update prompt}
\PromptSection{Plan generation instruction}
\begin{lstlisting}[style=promptstyle]
You are a task planning assistant. Given a current task and similar past experiences, generate a concise and actionable step-by-step plan.

Current Task: {QUERY}
Similar Past Experiences: {DISCRETE MEMORY}

Based on the current task and the similar experiences above, generate a step-by-step plan that:
1. Is concise and actionable
2. Breaks down the task into clear, sequential steps
3. Uses insights from similar experiences when relevant

Format your response as a numbered list of steps.
\end{lstlisting}

\PromptSection{Plan update instruction}
\begin{lstlisting}[style=promptstyle]
You are a task planning assistant. Update the original plan based on recent observations and actions.

Experience Memory: {DISCRETE MEMORY}
Original Plan: {PLAN}
Recent Screenshot Observations: {SCREENSHOTS}
Recent Actions: {ACTIONS}
Current Observation: {SCREENSHOT}

Based on the recent actions and observations, update the plan to:
1. Reflect what has been accomplished
2. Adjust remaining steps based on current progress
3. Keep it concise (3-5 steps total)
4. Clearly state the next single step. If the task is finished, yield STOP.
5. Do NOT repeat actions that have already failed.

Output format:
<plan>Your updated plan here</plan>
<subgoal>next single step</subgoal>
\end{lstlisting}
\end{PromptBox}

\clearpage
\begin{PromptBox}[colframe=DeepGreen, colback=LightGreenBg]{Action generation prompt}
\PromptSection[DeepGreen]{System instruction}
\begin{lstlisting}[style=promptstyle]
You are a helpful execution agent following the given plan.

Task: {QUERY}
Plan: {PLAN}
Available Actions: {ACTION_SPACE}

Constraints:
1. Follow the plan step by step.
2. Specify the element number to interact with.
3. Don't repeat failed actions.
4. Provide an answer within the remaining steps.
5. Output one action at a time.
\end{lstlisting}
\end{PromptBox}

\begin{PromptBox}[colframe=DeepOrange, colback=LightOrangeBg]{Memory check and update prompt}
\PromptSection[DeepOrange]{Input prompt}
\begin{lstlisting}[style=promptstyle]
You are helping a GUI agent decide if it should look for different reference experiences.

Original Task: {QUERY}
Recent Observations: {SCREENSHOTS}
Recent Actions: {ACTIONS}
Current Memories: {DISCRETE MEMORY}

Rules:
1. Memories are loose references - they don't need to match perfectly.
2. If current memories are related, output NO_UPDATE.
3. Only output NEEDS_UPDATE if the agent moved to a completely different activity type.

Output: "NO_UPDATE" or "NEEDS_UPDATE: <2-5 keywords>"
\end{lstlisting}
\end{PromptBox}

\clearpage
\begin{PromptBox}[colframe=DeepPurple, colback=LightPurpleBg]{VLM-as-Judge Evaluation Prompt}
\PromptSection[DeepPurple]{System instruction}
\begin{lstlisting}[style=promptstyle]
You are an expert at analyzing web browsing task completion from screenshots. You will be given a task instruction and a series of screenshots and corresponding plans of a web browsing session. 

Please analyze the screenshots and provide a detailed assessment of the task completion by following the steps below:

1. First, understand the task instruction. Describe what successful task completion should look like (i.e., what the final screenshot should show).
2. For each screenshot, analyze:
   - What is visible on the page (URL, elements, forms, buttons, messages, etc.)
   - What action the user performed (click, type, scroll, etc.)
   - Whether and how the action contributed to progress (or caused mistakes)
3. Carefully observe URL bars, form fields, search boxes, buttons, error messages, results, and confirmation screens.
4. After analyzing all screenshots, provide your reasoning about:
   - Whether the task was completed
   - How clear, logical, and efficient the interaction sequence was
   - Whether actions were redundant or irrelevant

Then, assign a score using this 3-level scale:

SCORING SCALE:

- 5 -- Task Fully Completed and Efficiently Done  
  - The final screenshot shows the task was clearly and completely fulfilled  
  - Actions followed a logical, goal-directed path  
  - No major mistakes or repetitions  
  - The agent shows strong understanding of the task

- 3 -- Task Partially Completed with Some Mistakes  
  - The task was partially fulfilled, but the final state is incomplete  
  - The agent made some progress, e.g., found the right page or filled part of a form  
  - There may be inefficient actions, repetition, or missed final steps  
  - The agent shows some understanding, but does not complete the task

- 1 -- Task Not Completed / Mostly Failed  
  - The task was not completed at all  
  - Actions are largely random, irrelevant, or repetitive  
  - The final screenshot does not reflect any meaningful progress  
  - The agent shows poor understanding of the goal

Your response must strictly follow this format:

TASK REQUIREMENT:  
[Summary of what the task is asking for and what success looks like]

SCREENSHOT ANALYSIS:  
Screenshot 1:  
[What is visible + what action was performed + how it contributed]  
Screenshot 2:  
...  
...

REASONING:  
[Overall reasoning -- was the task completed? Was the interaction logical? Any key mistakes?]

FINAL ANSWER:  
[Short conclusion about task success and agent behavior]

SCORE: [1/3/5]

Now, please strictly follow the format above and analyze the following screenshots:  
Task Instruction: {instruction}  
Screenshots (by order):
\end{lstlisting}
\end{PromptBox}

\end{document}